\documentclass[letterpaper, 10 pt, conference]{ieeeconf}
\IEEEoverridecommandlockouts
\usepackage[utf8]{inputenc}

\usepackage{caption}

\usepackage[style=ieee,url=false,doi=false,isbn=false]{biblatex}
\AtEveryBibitem{\clearfield{month}}
\AtEveryBibitem{\clearfield{day}}
\AtEveryBibitem{
    \clearfield{urlyear}
    \clearfield{urlmonth}
}
\usepackage{xcolor}
\usepackage{booktabs}
\usepackage{layouts}
\usepackage[notext]{stix}
\usepackage{amsmath}
\usepackage{amssymb}

\usepackage{mathtools}

\newcommand{\SEtwo}{\text{SE}(2)}
\newcommand{\powset}[1]{\mathcal{P}(#1)}

\DeclareMathOperator*{\argmax}{argmax}

\usepackage{siunitx}
\usepackage[hidelinks]{hyperref}

\usepackage[ruled,linesnumbered]{algorithm2e}
\SetAlgorithmName{Alg.}{alg.}{List of Algorithms}
\SetAlgoCaptionSeparator{:}
\SetAlCapSty{small}
\DontPrintSemicolon
\SetKwInput{KwInputs}{Inputs}
\SetKwInOut{KwOutputs}{Outputs}
\SetKw{KwFor}{for}
\SetKwData{Gets}{\ensuremath{\gets}}
\SetKwComment{Comment}{$\triangleright$\ }{}
\SetCommentSty{emph}

\usepackage[capitalise]{cleveref}
\crefname{figure}{Fig.}{Figs.}
\Crefname{figure}{Fig.}{Figs.}
\crefname{table}{Tab.}{Tabs.}
\Crefname{table}{Tab.}{Tabs.}
\crefname{section}{Sec.}{Secs.}
\crefname{Section}{Sec.}{Secs.}
\crefname{equation}{}{}
\Crefname{equation}{Equation}{Equations}
\crefname{algorithm}{Alg.}{Algs.}
\crefname{Algorithm}{Alg.}{Algs.}

\usepackage{tabularx}

\title{\LARGE \bf
Who Went Where When on the Lunar Surface:\\Forensic Trajectory Analysis to Identify Byzantine Rovers
}
\author{Lachlan Holden$^{1,2}$, Feras Dayoub$^{1}$, Melissa de Zwart$^{2}$, David Harvey$^{2}$, and Tat-Jun Chin$^{1,2}$
}


\begin{document}

\makeatletter
\twocolumn[{%
\begin{@twocolumnfalse}
\maketitle
\end{@twocolumnfalse}
}]
{
  \footnotetext{$^{1}$AI for Space Group and $^{2}$Andy Thomas Centre for Space Resources, Adelaide University, Australia.
{\tt\small \{lachlan.holden,\allowbreak{}feras.dayoub,\allowbreak{}melissa.dezwart,\allowbreak{}david.harvey,\allowbreak{}tat-jun.chin\}\allowbreak{}@adelaide.edu.au}}
}
\makeatother

\thispagestyle{empty}
\pagestyle{empty}

\begin{abstract}
Future planetary surface missions are likely to involve multiple independently operated rovers sharing the same deployment region, raising the need to verify compliance with operational constraints such as Lunar Safety Zones. Because continuous in-situ observability is rarely available, such verification requires post-hoc reconstruction of rover trajectories from sparse telemetry, including odometry, pose priors, and relative inter-rover detections. We introduce the problem of forensic trajectory analysis for non-cooperative planetary rovers in the presence of Byzantine agents: rovers that provide miscalibrated or deliberately falsified measurements to support an incorrect trajectory. We show that standard outlier-robust pose graph optimisation methods are vulnerable in this setting, because Byzantine rovers can generate measurements that are internally consistent and numerous enough to make truthful incriminating measurements appear as outliers. To address this, we propose an attribution-aware trajectory estimation method that reasons over rover credibility rather than individual measurement validity. The method evaluates candidate credible rover subsets by comparing the statistical consistency of their internal and boundary relative detections against provided priors, and then estimates trajectories using only measurements attributed to credible agents. Across synthetic simulations and real planetary-analogue trajectory data, the proposed method identifies Byzantine rovers and produces significantly more accurate trajectory estimates than existing robust pose graph optimisation baselines.
\end{abstract}


\section{Introduction}
One of the key characteristics of the New Space era of the 21st century is the rise of space missions driven by private companies \autocite{bousedra.2023_downstream}. With lunar landings no longer solely the domain of individual nation-states, current and future missions involve the rovers and landers of multiple companies and countries working together in close proximity. This is exemplified by missions like IM-2, which involved multiple US and Japan-based companies \autocite{intuitivemachines.2025_im}. For successful co-located operation of this kind, efforts like the Artemis Accords \autocite{NASA.2020_artemis} and the Chinese-led International Lunar Research Station \autocite{wu.2023_international} are attempting to establish norms. One relevant provision of the Artemis Accords is that of Lunar Safety Zones, where stakeholders can establish areas where only certain parties may enter.

To be able to determine the compliance of different rovers with protocols like Lunar Safety Zones under the situational awareness constraints typical of planetary missions, a method for post-hoc determining the trajectories that rovers have taken based on relatively low-fidelity downlinked data is essential. This data includes metrics like odometry, map localisation estimates, and relative detections from infrequent images. For the real-world application of these trajectories, it is crucial to design a method that is Byzantine-resilient, which is to say capable of handling agents that are providing poorly calibrated or deliberately misleading data. This work introduces this problem of ``forensic trajectory analysis'': post-hoc determination of multi-rover trajectories in the presence of adversarial measurements.

This paper makes three contributions: \begin{itemize}
    \item We formulate forensic trajectory analysis for non-cooperative planetary rovers as a post-hoc state-estimation problem with Byzantine measurement sources;
    \item We propose an attribution-aware credibility inference method that scores candidate rover subsets using internal consistency and boundary inconsistency of relative detections; and
    \item We demonstrate, using synthetic and planetary-analogue trajectories, that conventional robust pose graph optimisation (PGO) methods can be systematically misled by Byzantine agents, while the proposed method identifies non-credible rovers and improves trajectory reconstruction.
\end{itemize}

\section{Related work}
This work sits at the intersection of multi-robot localisation, robust PGO, and Byzantine-resilient robotics. Multi-robot localisation provides the estimation framework for fusing odometry, pose priors, and inter-robot observations. Robust PGO addresses false loop closures and other measurement outliers. Byzantine robotics studies unreliable or adversarial agents in distributed systems. Crucially, forensic trajectory analysis builds on and differs from each of these settings: our task is post-hoc reconstruction from attributed measurements, where entire rovers may act as a coherent but unreliable source of evidence.

\subsection{Multi-Robot Localisation}
One of the key problems in multi-agent robotic systems is that of localisation. While methods for accurate localisation of individual rovers over time exist \autocite{verma.etal.2024_enabling,kou.etal.2025_crosssite}, there has also been considerable investigation into how the relative detections between multiple rovers can be leveraged for even better estimation \autocite{chen.etal.2025_relative}.

These kinds of approaches are used as the basis for multi-rover collaborative localisation and mapping (C-SLAM), with key implementations DOOR-SLAM \autocite{lajoie.etal.2020_doorslam}, Kimera-Multi \autocite{tian.etal.2021_kimeramulti}, and Swarm-SLAM \autocite{lajoie.beltrame.2024_swarmslam}. Approaches for especially challenging environments like caves, mines, and planetary surfaces have been developed \autocite{ebadi.etal.2021_dareslam,chang.etal.2022_lamp,lajoie.etal.2025_multirobot}.

These systems generally assume that robots are cooperative participants in a shared estimation problem, whereas forensic trajectory analysis treats rovers as independently operated sources of evidence whose measurements may not be credible.

\subsection{Robust Pose Graph Optimisation}
Pose graph optimisation is a form of Bayesian state estimation that is commonly used for SLAM back-ends, especially for state-of-the-art multi-agent SLAM \autocite{lajoie.etal.2020_doorslam,tian.etal.2021_kimeramulti,ebadi.etal.2021_dareslam,lajoie.beltrame.2024_swarmslam}, with well-known forms for factors representing pose priors, odometry, relative measurements, and loop closures \autocite{dellaert.kaess.2017_factor}.

Perceptual aliasing causing loop closure outliers is a key challenge of SLAM. Two primary underlying approaches for outlier-robust PGO are Pairwise Consistency Maximisation (PCM) \autocite{mangelson.etal.2018_pairwise}, which underpins \autocite{lajoie.etal.2020_doorslam,ebadi.etal.2021_dareslam}, and Graduated Non-Convexity (GNC) \autocite{yang.etal.2020_graduated}, which underpins \autocite{tian.etal.2021_kimeramulti, lajoie.beltrame.2024_swarmslam,chang.etal.2022_lamp}. These are the two methods that we compare against in this paper.

PCM \autocite{mangelson.etal.2018_pairwise} and variations \autocite{olivastri.pretto.2024_ipc,forsgren.etal.2024_groupk} robustly merge two maps with loop closures between them by leveraging graph theory to efficiently find the maximum clique of pairwise-consistent measurements. GNC \autocite{yang.etal.2020_graduated} is a more general method for robust optimisation of any kind of factor graph, which uses robust cost functions with an iterative solver.

These robust PGO methods are very effective at handling perceptual aliasing and allowing successful SLAM back-end optimisation, but as we discuss further in \cref{sec:ours-vs-robust}, they are insufficient for identifying and rejecting actively misleading information from Byzantine agents.

\subsection{Byzantine Robotics}
While much of the literature focuses on the cases of cooperative agents working together to solve problems, there is also investigation into resilience against Byzantine agents -- those who are adversarial or misleading when participating in multi-agent systems. While Byzantine resilience is relevant to all manner of distributed and networked systems, within robotics it is frequently considered in the context of multi-agent consensus \autocite{luo.etal.2025_robotobchain,zhao.etal.2023_generic} or distributed learning \autocite{chen.etal.2023_byzantinerobust,li.etal.2022_byzantine}.

At the specific intersection of Byzantine resilience for robotic localisation, problems like Byzantine detection by comparing against GNSS signals in UAV swarms \autocite{hacohen.etal.2020_improved}, and target and formation tracking \autocite{cui.etal.2023_byzantine,gong.etal.2024_resilient} have been investigated. These works address challenges in an online filtering context and with local, constrained communication and connectivity. This contrasts against our post-hoc, forensic trajectory analysis focus in this paper, which motivates using PGO as our underlying method and hence the comparisons to PCM and GNC.

The closest work we are aware of is by \textcite{moroncelli.etal.2024_byzantine}, which also explicitly considers the problem of Byzantine agents in a SLAM or PGO-based setting. Our work contrasts against theirs in a few key ways: we consider more advanced Byzantine behaviour including groups of consistent, collaborative agents; we use a statistically-motivated criteria for identifying Byzantine agents; we explain and demonstrate the vulnerability of robust PGO to Byzantine attacks; and we include the attribution of measurements and the identification of Byzantine agents as a first-class output of our system.

Taken together, there is a clear gap in the literature: existing multi-robot localisation methods provide the estimation machinery, and robust PGO methods provide measurement-level outlier rejection, but neither addresses adversarial measurement sources that can generate coherent false trajectories. In contrast, Byzantine robotics studies unreliable agents, but rarely in a post-hoc pose-graph reconstruction setting with attributed odometry, pose priors, and relative detections. Our work addresses this gap by treating credibility as a property of rovers rather than individual measurements.

\begin{figure*}[tb] \centering
\includegraphics[width=0.85\textwidth]{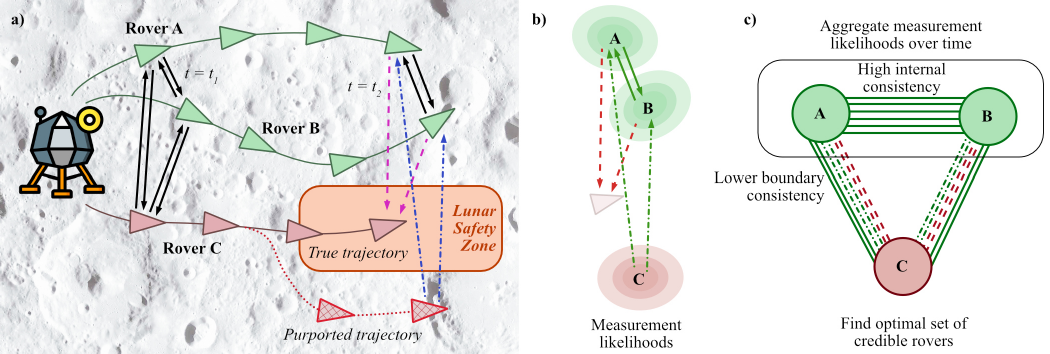}
\caption{A high-level overview of our method, showcasing an example of credible rovers A and B with a Byzantine rover C. \textbf{a)} The trajectories of the rovers and relative detections between them, including the purported trajectory by C intending to hide its incursion into the Lunar Safety Zone. \textbf{b)} The likelihoods of relative measurements at a given time compared to pose priors, where misleading measurements from C have high likelihood, while incriminating measurements from A and B have low likelihood. \textbf{c)} Finding the optimal set of credible rovers by determining the set with jointly highest internal consistency and lowest boundary consistency.} \label{fig:hero}
\end{figure*}

\section{Method}
In this section, we outline the problem formulation, our method for identifying non-credible rovers, and how we then calculate final trajectories. The process consists of first estimating putative trajectories from pose priors only; second scoring candidate credible rover subsets using internal and boundary consistency of relative detections; third selecting the credible set; and finally running PGO using only measurements attributed to credible rovers and weak smoothness constraints. This is illustrated further in \cref{fig:hero}. We then have a brief discussion of the fundamental limitations of robust PGO methods and how attribution of data is the key element that allows our method to work successfully in the Byzantine case.

\subsection{Problem Definition}
Each rover provides trajectory priors and relative detections, but some rovers may be non-credible sources. Our goal is to identify the credible sources and reconstruct all rover trajectories. We assume that at least one and no more than half of the rovers are non-credible.

Consider a collection of $|R|$ rovers
\begin{equation}
    R = \{ R_1, R_2, R_3, \ldots, R_{|R|} \} \text{.}
\end{equation}
If the state of a rover $R_r$ at time $t$ is denoted by
\begin{equation}
    x^{(r)}_t \in \SEtwo
\end{equation}
and the overall state of all rovers over all times $t = 0, \ldots, T$ is denoted by
\begin{equation}
    X = \{ x^{(r)}_t \mid r \in \{1, \ldots, |R|\}, t \in \{ 0, \ldots, T \} \} \text{,}
\end{equation}
we are interested in finding the optimal state estimate $X^\star$ that approximates the real, ground-truth states of the rovers $\tilde{X}$. In this work, we use $x^{(r)}_t \in \SEtwo$ as we assume pose priors and detections are reported relative to an overhead two-dimensional map, like those commonly used for on-board navigation.

We consider two kinds of measurements $z$, required to be provided by rover operators to achieve this goal -- pose priors, and relative detections of other rovers. It is reasonable to assume these measurements are available from rover operators, as they are readily derived from crucial localisation, navigation, and situational awareness systems. The pose priors have uncertainty at each timestep, and can take either an absolute form (like those from map-based localisation), or a relative form (like those from odometry.) Hence we have
\begin{equation}
    z_P \in Z_P \text{ where } Z_P = \{ z^{(r)}_{t} \text { or } z^{(r)}_{t-1,t}, \ldots \} \text{,}
\end{equation}
where the measurement takes the form of an absolute or relative pose $z_P \in \SEtwo$. For the relative detections of one rover by another at a particular timestep, we have
\begin{equation}
    z_D \in Z_D \text { where } Z_D = \{ z^{(r_1, r_2)}_t, \ldots \} \text{.}
\end{equation}
In this work, we consider three kinds of relative detections -- bearing only, where $z_D \in [\ang{0}, \ang{360})$, range and bearing, where $z_D \in \mathbb R \times [\ang{0}, \ang{360})$, and full degree-of-freedom range, bearing, and orientation, where $z_D \in \SEtwo$.

For some subset of rovers
\begin{equation}
    C \in \powset{R}
\end{equation}
where $\powset S$ denotes the power set of $S$, we can consider the relevant subset of measurements originating from rovers in $C$,
\begin{equation}\label{eq:ZC}
    Z^C = Z^C_P \cup Z^C_D = \{ z^{(r)}_{t} \mid R_r \in C\} \cup \{ z^{(r_1, r_2)}_t \mid R_{r_1} \in C\} \text{.}
\end{equation}

If we say that a rover $R_r$ is \emph{credible}, we take that to mean that each measurement it provides accurately represents the world state up to well-characterised noise, i.e.
\begin{equation}
    z^{(r)} \sim \mathcal{N}\left(h(\tilde{X}), \Sigma^{(r)} \right)
\end{equation}
for the relevant measurement function $h(X)$. If a certain subset of rovers $\tilde{C}$ are credible, and all other rovers $R \setminus\tilde{C}$ are non-credible, then our ultimate goal is to find the optimal state estimate $X^\star$ which based on only measurements from credible rovers $Z^{\tilde{C}}$. As the true set of credible rovers $\tilde{C}$ is impossible to know, the problem is realised in two steps where first an estimate of the credible rovers $C^\star$ is found, and then a state estimate $X^\star$ is made by considering $Z^{C^\star}$. This two step structure was found to perform more reliably than a joint optimisation in preliminary experiments.

\subsection{Identifying Byzantine Rovers}
We first consider putative trajectories $\hat{X}$ with uncertainties $\hat{\Sigma}$ for each rover based on the pose priors data only. These are found either taking $\hat{X}$, $\hat{\Sigma}$ from absolute priors directly, or by solving the following in the case where relative priors $Z_P$ are provided:
\begin{equation}\label{eq:getsigmas}
    \hat{X}, \hat{\Sigma} \leftarrow \argmax_{X} p(X \mid Z_P) \text{.}
\end{equation}

Then, for a candidate set of credible rovers $C \in \powset{R}$, we consider the set of \emph{internal} relative detections
\begin{equation}\label{eq:ZCI}
    Z^C_I = \{ z^{(r_1, r_2)}_t \mid z \in Z_D, R_{r_1} \in C, R_{r_2} \in C \}
\end{equation}
i.e. the measurements between two rovers in $C$, as well as the set of \emph{boundary} relative detections
\begin{equation}\label{eq:ZCB}
    Z^C_B = \{ z^{(r_1, r_2)}_t \mid z \in Z_D, R_{r_1} \in C \veebar R_{r_2} \in C \}
\end{equation}
where $\veebar$ represents exclusive OR. This is illustrated graphically in \cref{fig:hero}(c), with $C =\{R_A, R_B\}$. We are interested in determining the degree to which these measurements align with the provided priors -- more specifically, we want to determine how \emph{well} the internal set of measurements are consistent with the priors and how \emph{poorly} the boundary set of measurements are consistent with the priors. As such, we can then calculate the internal and boundary sets of residual errors for some error function $e(\cdot)$ as
\begin{align}
    E^C_I &= \{ e(z, \hat{X}, \hat{\Sigma}) \mid z \in Z^C_I \} \text{ and}\\
    E^C_B &= \{ e(z, \hat{X}, \hat{\Sigma}) \mid z \in Z^C_B \} \text{.}
\end{align}
If $|Z^C_I|=0$, the internal boundary set is degenerate and the candidate set $C$ is skipped.

To determine the form of $e(z, X, \Sigma)$ for the relative detection $z^{(r_1, r_2)}_t$, we first consider the raw measurement residual
\begin{equation}
    e_0 = z^{(r_1, r_2)}_t \ominus h(\hat{x}^{(r_1)}_t, \hat{x}^{(r_2)}_t)
\end{equation}
for measurement function $h(\cdot)$. The effective information matrix, taking into account the uncertainty of the relative detection itself $\Sigma_D \triangleq \hat{\Sigma}^{(r_1, r_2)}_t$ and the uncertainties of the two relevant poses $\Sigma_1 \triangleq \hat{\Sigma}^{(r_1)}_t$ and $\Sigma_2 \triangleq \hat{\Sigma}^{(r_2)}_t$, is
\begin{equation}
    \Lambda_\text{eff} = \Sigma_\text{eff}^{-1} = \left( \Sigma_D + H_1 \Sigma_1 H_1^\top + H_2 \Sigma_2 H_2^\top \right)^{-1}
\end{equation}
for Jacobians
\begin{equation}
    H_1 =
        \frac{\partial h}{\partial x_1}
        \text{ and }
    H_2 =
        \frac{\partial h}{\partial x_2}
       \end{equation}
evaluated at
\begin{equation}
x_1 = \hat{x}^{(r_1)}_t\text{ and } x_2 = \hat{x}^{(r_2)}_t \text{.}
\end{equation}
Interpreting the raw residual under these uncertainties therefore leads to the overall error function
\begin{equation}\label{eq:e}
    e(z, X, \Sigma) = \lVert z^{(r_1, r_2)}_t \ominus h(\hat{X}) \rVert^2_{\Sigma_\text{eff}} = e_0^\top \Lambda_\text{eff}\, e_0 \text{.}    
\end{equation}
This function has the nice property that as the pose prior uncertainties increase, a measurement that aligns less accurately with $\hat{X}$ will be punished less and receive a lower residual.

If the noise is properly characterised, it is determined that we would expect the residual to follow the distribution
\begin{equation}
    \lVert z - h(x) \rVert^2_\Sigma \sim \chi^2_\nu
\end{equation}
for $\nu$ degrees of freedom in the measurement (i.e. $\nu=1$ for bearing, $\nu=2$ for range-bearing, $\nu=3$ for full state). As such, we can determine the $p$-values indicating the probability with which, all rovers being credible, we would expect to see these internal and boundary residuals. Due to the additivity of the $\chi^2$ distribution, these can be calculated as
\begin{align} \label{eq:pCI}
    p^C_I &= P\left(\chi^2_{\nu\,\lvert r^C_I \rvert} \geq \textstyle\sum_{e \in E^C_I} e \right) \text { and } \\ \label{eq:pCB}
    p^C_B &= P\left(\chi^2_{\nu\,\lvert r^C_B \rvert} \geq \textstyle\sum_{e \in E^C_B} e \right) \text{.}
\end{align}

Finally, we can find a quality metric $q^C$ for a particular subset $C$ as the difference of log probabilities
\begin{equation} \label{eq:qC}
    q^C = \log p^C_I - \log p^C_B \text{.}
\end{equation}
Hence, our estimated set of credible rovers $C^\star$ can be found as
\begin{equation}\label{eq:Cstar}
    C^\star = \argmax_{C \in \powset{R}} q^C \text{.}
\end{equation}
Intuitively, we are interested in finding the group of rovers that simultaneously has the \emph{least} surprising relative measurements between each other and has the \emph{most} surprising relative measurements between themselves and out-group rovers.

Once we have our estimated set of credible rovers $C^\star$, we choose to trust the measurements attributed to these rovers, $Z^{C^\star}$. This set does not include any of the priors or relative detections that come from the deemed non-credible rovers $R \setminus C^\star$. Discounting the priors completely in this way causes the pose graph to become under constrained, so different constraints need to be added. The laws of physics dictate that even for a non-credible rover, the state $x^{(r)}_{t+1}$ is going to be within some neighbourhood of $x^{(r)}_{t}$, represented by an expected relative pose offset
\begin{equation}
    \breve{z} \in \SEtwo = [\delta_\text{forwards}, 0, 0]
\end{equation}
which indicates expected forwards motion and a corresponding, considerably large uncertainty in all three components. As such, we supplement our observations with ``smoothness constraints''
\begin{equation}\label{eq:Zbreve}
    \breve{Z} = \{\breve{z}_{t,t+1}^{(r)} \mid r \in R \setminus C^\star, t \in \{0,\ldots,T-1\}\}
\end{equation}

We can then finally use typical pose-graph optimisation via the Levenberg-Marquardt method to estimate the trajectories of all rovers $X^\star$, which optimises:
\begin{equation}\label{eq:Xstar}
    X^\star = \argmax_{X} p\left(X \mid Z^{C^\star}, \breve{Z} \right) \text{.}
\end{equation}

An overview of this method is shown in \cref{alg:overview}.

\begin{algorithm}
\caption{Algorithmic overview of the method.}\label{alg:overview}
\KwInputs{Pose priors $Z_P$ and relative detections $Z_D$}
\KwOutputs{Trajectory estimates $X^\star$}
Get $\hat{X}$, $\hat{\Sigma}$ from $Z_P$ \Comment*[r]{pose prior estimates \cref{eq:getsigmas}}
\For{$C \in \powset{R}$}{
    Get $Z^C_I$, $Z^C_B$ from $Z_D$ \Comment*[r]{\cref{eq:ZCI}, \cref{eq:ZCB}}
    Get $\;\;E^C_I$, $E^C_B$ from $e(\cdot)$ on $Z^C_I$, $Z^C_B$ \Comment*[r]{\cref{eq:e}}
    Get $p^C_I$, $p^C_B$ from $E^C_I$, $E^C_B$ \Comment*[r]{\cref{eq:pCI}, \cref{eq:pCB}}
    $q^C \;\Gets\; \log{p^C_I} - \log{p^C_B}$ \Comment*[r]{quality score \cref{eq:qC}}
}
$C^\star \gets \argmax_{C \in \powset{R}} q^C$ \Comment*[r]{most credible set \cref{eq:Cstar}}
Get $Z^{C^\star}$ and $\breve{Z}$ \Comment*[r]{credible \cref{eq:ZC} \& smoothing \cref{eq:Zbreve}}
$X^\star \Gets \argmax_X p( X \mid Z^{C^\star}, \breve{Z})$ \Comment*[r]{LM optimiser \cref{eq:Xstar}}
\KwRet $X^\star$
\end{algorithm}

\subsection{Contrasting Against Robust Methods}\label{sec:ours-vs-robust}
The fundamental difference between our method and robust PGO methods like GNC and PCM is that we consider the \emph{attribution} of measurements. Robust PGO methods are designed to mitigate false positive loop closures in SLAM, and as such have an underlying assumption that the true set of inlier measurements will be internally consistent and any consistent set of outliers will be smaller than the true set. This is encapsulated in their overall optimisation formulation
\begin{equation}
    X^\star = \argmax_X p\left( X \middle| Z \right)
\end{equation}
which includes \emph{all} measurements $Z$.

In the Byzantine case, this assumption no longer holds -- Byzantine agents can produce essentially arbitrary volumes of data that is internally consistent and consistent with the true world state of credible agents, while being misleading about their own state. As an illustrative example, consider \cref{fig:hero}(a), where believing rover C's false detections and consistent priors (four measurements in total) only requires disbelieving two true, incriminating edges. Maximising the likelihood over \emph{all} measurements is going to prefer the solution which disbelieves fewer measurements. Thus, treating the Byzantine identification problem as a problem of rejecting outlier measurements fails systematically under coherent Byzantine attacks, and this is supported experimentally in the following section.

Furthermore, while GNC implicitly allows for outliers in odometry or pose prior measurements, PCM explicitly only identifies relative detections as outliers when they disagree with priors and so will always side with misleading measurements over incriminating ones. Considering the attribution of measurements, as our method does, is crucial in the Byzantine case.

\section{Experiments}
In this section, we evaluate the performance of our method against both simulated and real planetary-analogue trajectory data. We compare against GNC and PCM as representatives of a robust PGO approach, to demonstrate that this class of method -- designed for robustness to false positive loop closures -- is fundamentally untenable in the Byzantine case.

\subsection{Simulation Definition}
We begin by demonstrating our method in a 2D simulation environment. We consider groups of $|R|=5$ rovers, and evaluate across five different initial positions.

\subsubsection{Motion Model}
Each rover is modelled as a white noise on acceleration model, and hence is parameterised as
\begin{equation}
  R_r = (x_0, v_0, \omega_0, \sigma_v, \sigma_\omega)^{(r)} \text{,}
\end{equation}
with update equations
\begin{align}
    v_t &\sim \mathcal{N}(v_{t-1}, \sigma_v^2) \\
    \omega_t &\sim \mathcal{N}(\omega_{t-1}, \sigma_\omega^2) \\
    x_t &= x_{t-1} + \begin{bmatrix}
        v_t\cos(\theta_{t-1}+\omega_t) \\
        v_t \sin(\theta_{t-1} + \omega_t) \\
        \omega_t
    \end{bmatrix} \text{.}
\end{align}
We use parameters $v_0 = \SI{1}{\meter\per\second}$, $\sigma_v = \SI{0.02}{\meter\per\second}$, $\omega_0 = \SI{0}{\per\second}$ and $\sigma_\omega = \SI{0.2}{\per\second}$, which produce relatively smooth trajectories representative of rover traverses.

\subsubsection{Trajectory Priors}
We consider two modes of supplied trajectory priors -- relative and absolute -- both created by perturbing the true trajectory with Gaussian noise. This is somewhat idealised, but serves to demonstrate the capability of our method independent of front-end performance.

For the former, relative case, we use a noise model of $\sigma_\text{forward} = \SI{0.3}{\meter}$, $\sigma_\text{left} = \SI{0.2}{\meter}$, $\sigma_\theta = \ang{2}$ to perturb the true relative pose between subsequent timesteps. A single pose prior at $t=0$ is also used to ground the odometry in the world frame, with noise model $\sigma_x=\sigma_y=\SI{2}{\meter}$, $\sigma_\theta = \ang{5}$.

For the latter, absolute, case, we use a noise model of $\sigma_x=\sigma_y=\SI{1}{\meter}$, $\sigma_\theta = \ang{5}$ to independently perturb the true absolute pose at each timestep.

For the smoothness constraints $\breve{z}$ we use $\delta_\text{forward} = \SI{1}{\meter}$, $\sigma_\text{forward}=\SI{0.5}{\meter}$, $\sigma_\text{left}=\SI{0.25}{\meter}$, and $\sigma_\text{turn}=\ang{5}$.

\subsubsection{Relative Detections}
For the relative detections between rovers, we use a noise model of $\sigma_\text{range}=\SI{3}{\meter}$, $\sigma_\text{bearing}=\ang{5}$, and $\sigma_\text{orientation}=\ang{5}$. Each rover only produces a relative observation of other rovers every $t=5$ timesteps, and only observes rovers within a certain field-of-view of its forwards and backwards directions. We evaluate the effect of differing fields of view in the following section.

\subsubsection{Byzantine Behaviour}
We consider the effects of both one and two Byzantine agents on our group of five rovers. We leverage the concept of a ``purported trajectory'' to define the Byzantine behaviour, where this is the path that they intend for us to believe was taken. All measurements sourced from a Byzantine agent are based on perturbed (as above) measurements from this purported trajectory. Credible rovers still provide relative measurements to the Byzantine rover's true state, as expected. Additionally, in the case of two Byzantine agents, we assume they are collaborating; i.e. the relative detections they provide \emph{between} one another correspond to the differences between their respective purported trajectories and hence are internally consistent.

We consider a range of classes of purported trajectories. All are based on transformations of the rover's initial trajectory, with different combinations of distance scaling, and offsets in translation and rotation applied. We additionally consider a ``late turn'', where the rover's purported trajectory aligns with its true trajectory for the first half of the experiment, and then diverges for the second half.

\subsection{Simulation Experiments}
The simulation environment is used for a series of comparisons against GNC and PCM. The most reasonable relative measurements to expect in a non-cooperative planetary context are either bearing (B) or range-bearing (RB), so we use these degrees of freedom to evaluate our method and GNC. The full three degree-of-freedom range-bearing-orientation (RBO) measurements are required for PCM, so we evaluate that method using these. We use $N=5$ runs with varied trajectories for each individual configuration of parameters.

\subsubsection{One Byzantine Rover}
\begin{figure}
    \centering
    \includegraphics[width=\columnwidth]{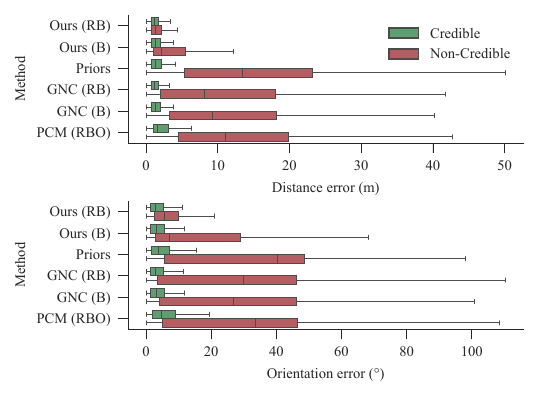}
    \caption{Distance and orientation errors at each timestep for credible and non-credible rovers by method in the case of a single Byzantine rover.}
    \label{fig:sim-errors-one}
\end{figure}

\begin{figure}
    \centering
    \includegraphics[width=\columnwidth]{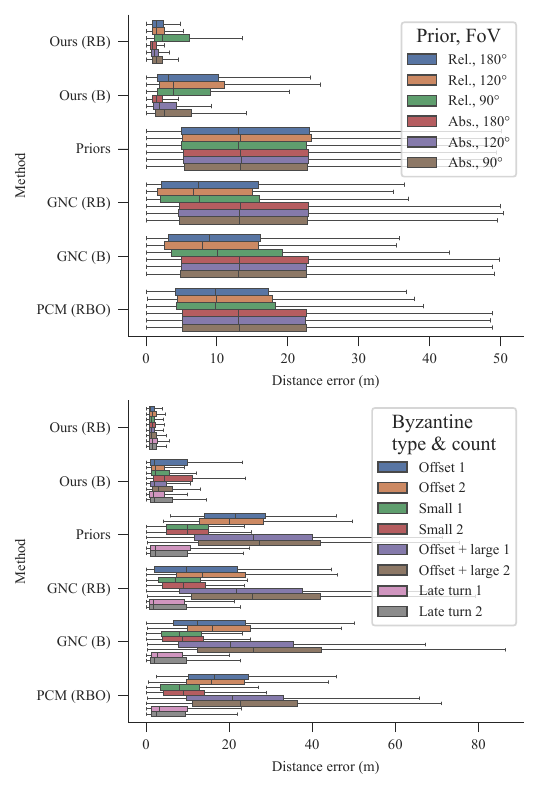}
    \caption{A breakdown of distance errors at each timestep of non-credible rovers only by prior measurement kind \{relative, absolute\}, front and rear field of view (top), and by type of Byzantine behaviour and number of Byzantine agents (bottom).}
    \label{fig:sim-errors-breakdown}
\end{figure}

First, we consider the cases of a single Byzantine rover, relative and absolute trajectory priors, and front and rear fields of view of \ang{180}, \ang{120}, and \ang{90}. As demonstrated in \cref{fig:sim-errors-one}, while all methods provide relatively accurate localisation results for the credible rovers, only our method provides at all comparable localisation results for the Byzantine rovers across all of the cases under consideration. This improved performance is seen across all measurement kinds, fields of view, and types of Byzantine behaviour, as shown in \cref{fig:sim-errors-breakdown}.

This can be explained by considering the recall metric in \cref{tab:sim-recall}, where we evaluate the proportion of non-credible measurements successfully rejected as outliers by each method. Our method of identifying non-credible rovers and rejecting all of their measurements performs very well, as indicated by the 83.3\% and 90.8\% recalls, while the robust PGO methods are misled by the internally consistent Byzantine measurements, as indicated by recalls of less than 10\%. These methods' lower precision in \cref{tab:sim-recall} further indicate that outlier rejection does not accurately detect Byzantine measurements.

\begin{table}
    \centering
    \caption{Recall and precision of non-credible measurements by method, detection model, and number of Byzantine agents.}
    \label{tab:sim-recall}
    \begin{tabular}{@{}llrrrrrr@{}}
\toprule
Method & Model & \multicolumn{3}{c}{Recall (\%)} & \multicolumn{3}{c}{Precision (\%)} \\ \cmidrule(rl){3-5} \cmidrule(l){6-8}
& & 1 Bz. & 2 Bz. & Total & 1 Bz. & 2 Bz. & Total \\ \midrule
Ours & RB & 90.8 & 89.2 & 90.0 & 86.7 & 94.2 & 90.4 \\
& B & 83.3 & 64.6 & 74.0 & 80.8 & 71.7 & 76.2 \\
GNC & RB & 7.1 & 4.4 & 5.7 & 20.1 & 13.5 & 16.8 \\
& B & 4.2 & 5.0 & 4.6 & 19.8 & 18.3 & 19.0 \\
PCM & RBO & 6.4 & 5.4 & 5.9 & 31.1 & 45.2 & 38.1 \\
\bottomrule
\end{tabular}
\end{table}

\begin{figure}
    \centering
    \includegraphics[width=\columnwidth]{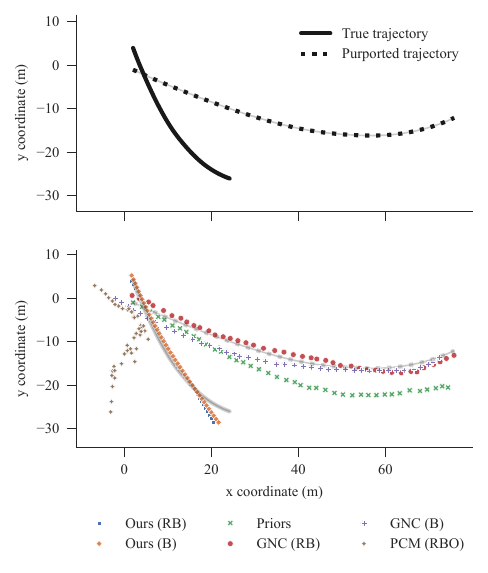}
    \caption{Qualitative comparison of true trajectory, purported trajectory, and trajectory results for a single Byzantine rover in one of the simulation runs.}
    \label{fig:sim-qualitative}
\end{figure}

This is further illustrated by the qualitative trajectories compared in \cref{fig:sim-qualitative}, where the results from GNC largely follow the Byzantine rover's \emph{purported} trajectory and PCM fails to produce to a meaningful estimate, while ours successfully leverages the data from the credible rovers to track its true trajectory instead.

\subsubsection{Two Byzantine Rovers}
\begin{figure}
    \centering
    \includegraphics[width=\columnwidth]{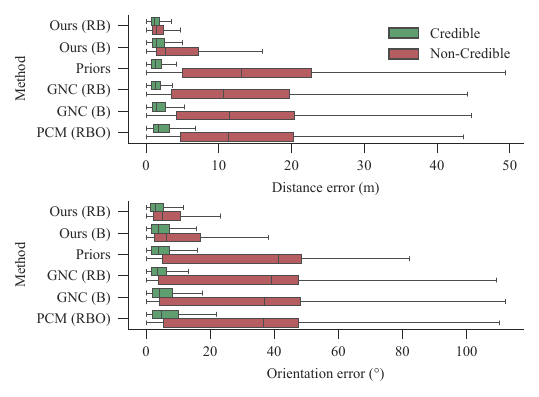}
    \caption{Distance and orientation errors at each timestep for credible and non-credible rovers by method in the case of two Byzantine rovers.}
    \label{fig:sim-errors-two}
\end{figure}

Largely the same trend is demonstrated for the case of two collaborative Byzantine rovers, as indicated in \cref{tab:sim-recall} and \cref{fig:sim-errors-two}. This again is consistent across experimental conditions as demonstrated in \cref{fig:sim-errors-breakdown}. The recall performance in the case of bearing-only measurements under the case of two Byzantine rovers is markedly poorer than other circumstances, but still outperforms robust PGO by achieving 64.6\% compared to their approximately 5\% each.

\subsubsection{Simulated Occlusions}
\begin{figure}
    \centering
    \includegraphics[width=\columnwidth]{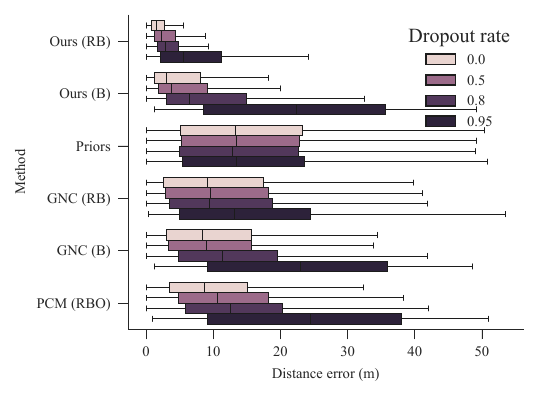}
    \caption{Distance errors at each timestep for non-credible rovers by method in the case of increasing relative measurement dropout rates.}
    \label{fig:sim-dropout}
\end{figure}

To investigate the cases where rovers may not have clear views of one another, we run another suite of experiments where relative measurements have a probability of not being captured (``dropout'') at each potential timestep, rather than being constrained by rover field-of-view. As shown in \cref{fig:sim-dropout,tab:supp-recall}, our method maintains relatively consistent recall and precision performance up to even 80\% dropout. The same trend as above, where bearing-only measurements perform worse than range-bearing measurements, is seen.

\begin{table}
    \centering
    \caption{Recall and precision by method, detection model, and dropout rate.}
    \label{tab:supp-recall}
    \scriptsize
    \begin{tabular}{@{}llrrrrrrrr@{}}
\toprule
Method & Model & \multicolumn{4}{c}{Recall (\%)} & \multicolumn{4}{c@{}}{Precision (\%)} \\ \cmidrule(lr){3-6} \cmidrule(l){7-10}
& \multicolumn{1}{r}{\emph{Drop.:}} & 0.0 & 0.5 & 0.8 & 0.95 & 0.0 & 0.5 & 0.8 & 0.95 \\ \midrule
Ours & RB & 83.8 & 81.2 & 81.2 & 66.7 & 85.0 & 81.2 & 85.0 & 58.8 \\
& B & 77.5 & 68.8 & 67.9 & 50.0 & 85.0 & 73.8 & 71.8 & 50.0 \\
GNC & RB & 8.5 & 4.3 & 2.1 & 0.2 & 27.5 & 18.6 & 18.5 & 2.5 \\
& B & 5.7 & 7.6 & 1.5 & 0.0 & 33.0 & 32.4 & 14.5 & 0.0 \\
PCM & RBO & 7.6 & 4.6 & 6.3 & 16.7 & 30.3 & 26.4 & 20.6 & 20.0 \\
\bottomrule
\end{tabular}
\end{table}

\begin{figure}\centering
    \includegraphics[width=\columnwidth]{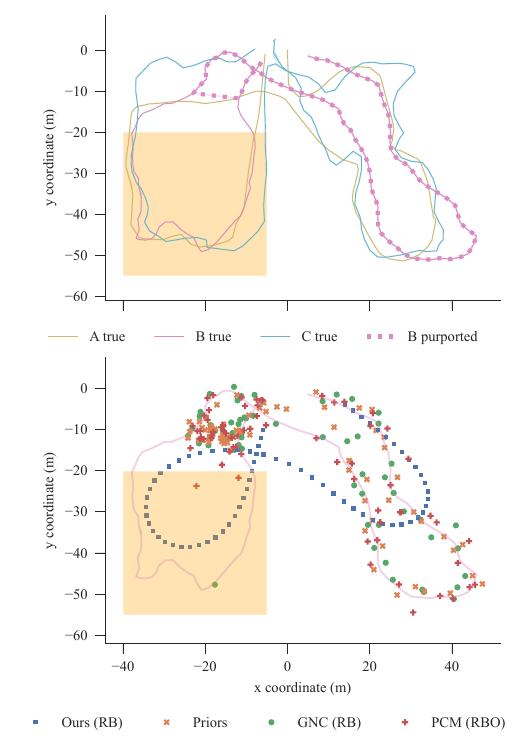}
    \caption{The trajectories of three rovers A, B, and C, and the purported trajectory of Byzantine rover B avoiding the Safety Zone marked in orange (top); and a comparison of the trajectory estimates produced by different methods (bottom).}
    \label{fig:gps-qualitative}
\end{figure}

\subsection{Planetary Analogue Experiments}
To further validate the concept, the C-SLAM dataset collected by \textcite{lajoie.etal.2025_multirobot} is used, which has RTK GPS data serving as ground truth trajectories for three rovers exploring a planetary analogue environment. We consider the scenario where there is a Safety Zone that rover~B is prohibited from entering, and so rover~B purports a ``shortcut'' trajectory that bypasses the time it spent within the zone, as indicated in \cref{fig:gps-qualitative}~(top).

For this experiment, we take the absolute poses at \SI{0.3}{\hertz} and perturb them with noise model $\sigma_x = \sigma_y = \SI{2}{\meter}$, $\sigma_\theta = \ang{5}$, which is more than generous based on the results in \autocite{verma.etal.2024_enabling}. We use a relative detection model with $\sigma_\text{range}=\SI{1}{\meter}$, $\sigma_\text{bearing}=\ang{3}$ (plus $\sigma_\text{orientation}=\ang{3}$ for PCM), with detections captured every $t=8$ timesteps, corresponding to almost \SI{30}{\second} of real time.

Our method successfully identifies rover~B as Byzantine. Furthermore, consider the plotted final trajectory estimates in \cref{fig:gps-qualitative}~(bottom). Although considerably smoothed due to the relative infrequency of relative measurements from credible rovers to guide rover~B's trajectory, our method is the only one that clearly and unambiguously shows that rover~B intrudes into the Safety Zone for a considerable amount of time, which is one of the most important outcomes for this problem of forensic trajectory analysis.

\section{Conclusions}
In this paper, we have introduced the problem of forensic trajectory analysis grounded in the needs of co-located operation of rovers by multiple stakeholders simultaneously. We demonstrate that traditional robust PGO is unable to provide suitable trajectory estimates in the case of Byzantine agents, and have introduced a statistically-motivated novel method based on the consistency of relative measurements observation and between groups of rovers. Experimental analysis has validated the accuracy of this method across a range of synthetic and real planetary analogue trajectories.

Future work should include investigating performance when using data from real odometry, localisation, and relative detection front-ends. It could also include a greater variety of Byzantine behaviours, including multiple disjoint groups of collaborative rovers, as well as further validation on real space or planetary analogue robotic platforms.

\AtNextBibliography{\small}
\printbibliography

\end{document}